# SeaAlert: Robust Severity Classification and LLM-Based Information Extraction for Noisy Maritime Distress Communications

**Tomer Atia[1], Yehudit Aperstein[2], and Alexander Apartsin[1]**
[1]School of Computer Science, Faculty of Sciences, HIT-Holon Institute of Technology, Holon 58102, Israel
[2]Intelligent Systems, Afeka Academic College of Engineering, Tel Aviv 69988, Israel

**ABSTRACT** Maritime distress communications transmitted over very high frequency (VHF) radio are safety-critical voice messages used to report emergencies at sea. Under the Global Maritime Distress and Safety System (GMDSS), such messages follow standardized procedures and are expected to convey essential details, including vessel identity, position, nature of the distress, and required assistance. In practice, however, automatic analysis remains difficult because distress messages are often brief, noisy, and produced under stress, may deviate from the prescribed format, and are further degraded by automatic speech recognition (ASR) errors caused by channel noise and speaker stress. This paper presents SeaAlert, a controlled experimental framework for evaluating robust analysis of maritime distress communications using transformer-based severity classification and LLM-based structured extraction. To address the scarcity of labeled real-world data, we develop a synthetic data generation pipeline in which an LLM produces diverse maritime messages, including challenging variants in which standard distress codewords are omitted or replaced with less explicit expressions. The generated utterances are synthesized into speech, degraded with simulated VHF noise, and transcribed by an ASR system to obtain controlled noise-degraded transcripts. The resulting evaluation shows that transformer-based classification degrades more gracefully than lexical baselines under ASR noise and codeword masking, while LLM-based extraction is more effective than Regex-based extraction for noisy structured fields.

## I. INTRODUCTION

Maritime safety systems rely heavily on the accurate and timely classification of distress signals to coordinate effective search-and-rescue operations. The Global Maritime Distress and Safety System (GMDSS) provides a standardized framework for emergency communications, yet its effectiveness depends on the ability to correctly interpret messages under challenging conditions [1]. Traditional Keyword Spotting (KWS) approaches, which detect predefined codewords, may be brittle in real-world scenarios due to noise, ambiguous phrasing, and contextual complexity [2]. These limitations can lead to delayed responses or false alarms, with potentially life-threatening consequences.

Recent advances in natural language processing (NLP) offer promising solutions to these challenges. Transformer-based models, such as RoBERTa, have demonstrated remarkable capabilities for understanding context and handling noisy inputs across various domains. However, their application to maritime communication remains underexplored, particularly compared to traditional methods such as Bag-of-Words (BoW) and rule-based codeword detection. Maritime environments present unique challenges, including radio interference, non-standardized phrasing, and the critical need for real-time processing.

This study addresses these gaps by developing and evaluating a controlled experimental pipeline for maritime distress communication analysis that combines severity classification with structured information extraction under simulated ASR noise. The pipeline is designed to isolate how different modeling choices behave under controlled variations in message severity, GMDSS codeword use, synthetic channel degradation, and adversarial wording. We examine whether transformer architectures provide greater robustness than BoW methods, particularly in noisy and adversarial conditions, due to their superior contextual understanding and robustness to variations in input quality. Our work builds on existing maritime safety research [3] while introducing a reproducible evaluation setting for studying robustness before larger-scale validation on operational recordings.

The primary contributions of this work are as follows. First, we present an end-to-end controlled evaluation framework for maritime distress communication analysis that combines synthetic message generation, VHF-style noise simulation, ASR transcription, severity classification, and downstream structured information extraction. Second, we provide a robustness-oriented comparison between classical Bag-of-Words baselines, a rule-based GMDSS keyword baseline, and transformer-based models, showing that although performance on clean text is broadly comparable, RoBERTa degrades less severely under ASR corruption, codeword masking, and adversarial message formulations within the evaluated synthetic setting. Third, we show that both model families rely strongly on explicit GMDSS codewords, but that RoBERTa generalizes more effectively when these lexical anchors are removed. Finally, we demonstrate that GPT-4o-mini-based extraction outperforms Regex-based extraction when evaluated against reference metadata fields defined during dataset construction, especially for alphanumeric and semantically variable fields. Together, these findings provide practical guidance for future maritime communication intelligence systems, while underscoring the need for validation on real VHF traffic before operational deployment.

To support reproducibility, the data, code, and implementation resources for SeaAlert are publicly available at: https://github.com/Tomeratia/SeaAlert.

The remainder of this paper is organized as follows: Section 2 reviews related work in maritime safety systems and text classification techniques. Section 3 details our methodology, including dataset construction and experimental design. Section 4 presents the comparative results across various noise levels and adversarial conditions. Section 5 discusses the implications of our findings for maritime safety applications, and Section 6 concludes with directions for future research.

## II. PRIOR WORK

### A. MARITIME EMERGENCY COMMUNICATION AND SAFETY CONTEXT

Maritime emergency response operates within a highly structured but operationally challenging communication environment. Foundational work on maritime safety systems has described the role of standardized procedures and communication protocols in search-and-rescue coordination [3], while later studies emphasized that the effectiveness of emergency response depends not only on the existence of such protocols, but also on accurate information sharing, timely coordination, and reliable interpretation of incident reports [1], [4], [5]. Research on shipboard communication failures further showed that communication breakdowns in maritime emergencies can directly increase operational risk and response severity [2]. Related studies on the human factor in maritime safety and on communication between ship and shore highlighted the practical difficulty of maintaining clear, context-aware interaction under stress, incomplete information, and noisy operational settings [6], [7]. Surveys of maritime communication technologies also noted that real-world maritime channels remain constrained by interference, bandwidth limitations, and heterogeneous communication infrastructures, all of which complicate robust automated interpretation [8], [9].

This literature establishes that maritime emergency communication is both safety-critical and operationally fragile. However, most of this body of work focuses on system design, risk factors, coordination processes, or human and organizational aspects, rather than on automatic understanding of short distress messages as noisy text inputs. As a result, the specific problem of NLP-based severity assessment from maritime VHF-style distress communication remains only weakly covered.

### B. NLP FOR MARITIME AND SAFETY-RELATED TEXT ANALYSIS

Text classification has long been studied in NLP, with classical methods such as bag-of-words, TF-IDF, linear models, and probabilistic classifiers remaining strong baselines in many applied settings [10]. More recently, transformer architectures have reshaped text understanding by enabling contextual token representations and stronger sequence modeling [11], [12], with review studies confirming their broad effectiveness across classification tasks [13]. In parallel, the automated processing of unstructured operational documents has become an important line of research in applied AI [14].

Within maritime and safety domains, prior NLP work is still relatively limited. Early studies explored the application of NLP to maritime communications [15], but these efforts predate modern transformer-based methods and do not address contemporary robustness challenges such as ASR corruption or adversarial phrasing. More recent work has applied NLP and deep learning to maritime risk assessment [16], while broader reviews of NLP on safety occurrence reports have shown growing interest in extracting actionable knowledge from safety-related textual records [17]. These studies demonstrate that safety-oriented text analytics is feasible and valuable, but they largely focus on post hoc reports, structured incident documentation, or risk assessment narratives rather than on short, time-critical radio messages transmitted in real time.

This distinction is important. Distress voice communications differ substantially from safety reports or routine documents: they are brief, formulaic but variable, often incomplete, and highly sensitive to transcription errors. Therefore, results from general text classification or from safety-report mining cannot be directly assumed to transfer to maritime distress communication.

### C. ROBUSTNESS OF NLP MODELS UNDER NOISE, AMBIGUITY AND ADVERSARIAL FORMULATIONS

Robustness is especially important in safety-critical AI systems, where model errors can have operational consequences [18], [19], [20]. Although transformer models generally outperform earlier architectures in many NLP tasks [11], [12], their behavior under noisy or adversarial inputs remains an active concern. Prior work has shown that popular transformer-based NLP models are vulnerable to noisy text perturbations [21], and that improving robustness requires dedicated architectural or training interventions rather than relying on standard pretraining alone [22], [23]. Concerns have also been raised regarding the interpretability and stability of neural text classifiers in high-stakes settings [24].

These issues are particularly relevant for maritime radio communication. In this domain, automatic speech recognition errors can corrupt critical lexical cues, while operational wording may include negations, speculation, drills, relays, or ambiguous procedural language. Existing NLP research has shown that negation and scope resolution remain challenging even for transformer-based architectures [25], [26], and noisy-domain entity recognition likewise becomes significantly harder when surface forms are degraded [27]. From an operational perspective, this means that recognizing explicit emergency codewords may not be sufficient; models must also reason correctly when those codewords are omitted, distorted, or contextually negated.

Despite these insights, prior maritime NLP studies have not systematically examined robustness to ASR-induced corruption, codeword masking, or adversarial trap messages in a unified evaluation framework. This remains a major gap for practical deployment in maritime safety applications.

### D. INFORMATION EXTRACTION, SYNTHETIC DATA, AND LLM-BASED PROCESSING

Beyond message-level classification, maritime response workflows require extraction of actionable fields such as vessel identifiers, position, persons on board, and nature of incident. Hierarchical and multi-task information extraction methods have shown that structured extraction can benefit from richer modeling of interdependent fields [28]. At the same time, work on knowledge-enhanced neural systems has highlighted the value of integrating contextual and logical constraints into language understanding pipelines [29]. However, extraction in noisy operational domains remains difficult, especially for alphanumeric identifiers and semantically variable slot values [27].

Data scarcity is another major limitation in safety-critical NLP. Large, labeled collections of real maritime distress communications are difficult to obtain, which motivates the use of synthetic data and augmentation. Surveys on text data augmentation have documented a broad range of strategies for improving robustness and coverage in low-resource classification tasks [30], while recent work on synthetic text generation with large language models showed both the promise and the limitations of LLM-generated training data [31]. In parallel, large language models have demonstrated strong few-shot and instruction-following capabilities [32], with scaling studies suggesting that larger models can better support generalization across language tasks [33]. These developments have encouraged growing interest in using LLMs in transportation and safety-oriented systems [34].

Recent work also emphasizes that LLM-generated synthetic data should be treated as a controlled experimental resource rather than as a direct substitute for real-world data. Synthetic data generated by LLMs can support text classification, but its effectiveness varies across tasks and is sensitive to task subjectivity [31]. Benchmark-oriented studies further show that synthetic benchmarks may approximate real-data evaluation in some NLP tasks, but their representativeness is task-dependent and may be affected by generator-related bias [35]. More recent work on evaluating language models as synthetic data generators also argues for standardized settings and metrics, because a model’s ability to solve a task does not necessarily imply that it is the best generator of training or benchmark data [36]. Controlled synthetic-benchmark construction has also been used in semantic text-relation evaluation, where LLM-generated examples are filtered to preserve intended information structure [37]. To the best of our knowledge, standardized public benchmarks and evaluation protocols for noisy maritime VHF/ASR distress transcripts remain limited. This gap motivates the controlled interpretation adopted in the present study: SeaAlert is used as a synthetic benchmark for robustness analysis, while validation on real maritime traffic and human-verified operational annotations remains necessary for deployment-level claims.

Against this background, the present study focuses on the specific setting of maritime distress communication, where both classification and structured extraction must remain reliable under ASR noise and protocol variation. In particular, prior work does not provide a clear comparison between rule-based extraction and LLM-based extraction for noisy maritime radio transcripts, nor does it connect such extraction performance to upstream robustness in severity classification.

### E. RESEARCH GAP

Prior work provides strong foundations in four relevant areas: maritime emergency communication and coordination [1-5], [7-9]; NLP-based text classification and transformer modeling [10], [11-13]; robustness challenges in noisy and adversarial language understanding [18], [21-24], [25-27]; and structured extraction and synthetic data generation for low-resource text analytics [14], [28-31]. However, these strands have not yet been integrated into a single end-to-end evaluation framework for maritime distress communication.

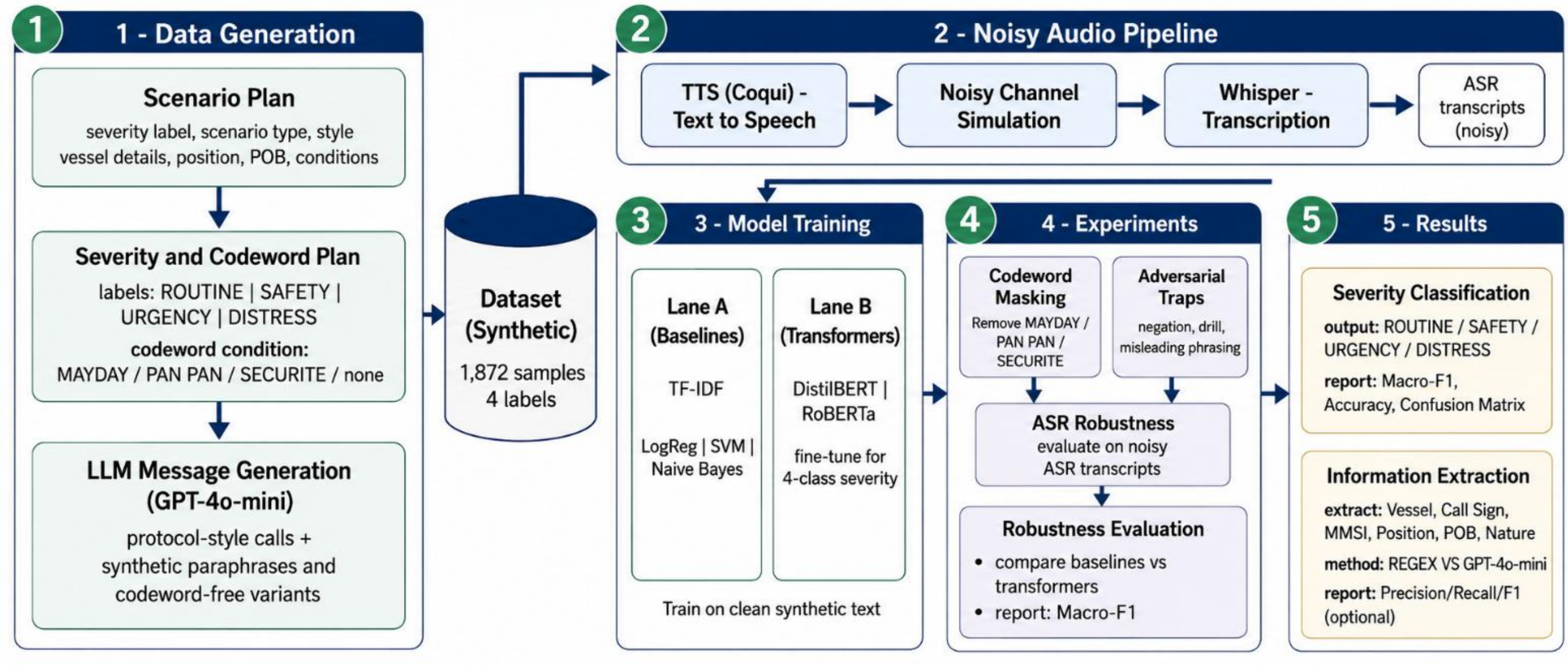

**FIGURE 1.** **SeaAlert experimental pipeline for synthetic data generation, noisy ASR processing, model training, robustness testing, and evaluation.**

The present study addresses this gap by examining maritime distress message understanding as a joint problem of severity classification and structured information extraction under simulated ASR corruption. Unlike prior maritime NLP efforts, the study explicitly compares classical bag-of-words and transformer-based classifiers, evaluates robustness under noise, codeword masking, and adversarial formulations, and contrasts Regex-based extraction with LLM-based extraction on noisy transcripts. This positioning distinguishes the work from both general text classification studies and broader maritime safety literature, while aligning it with the practical requirements of safety-critical communication intelligence.

## III. METHODOLOGY

The experimental framework for evaluating maritime distress signal classification consists of five interconnected stages designed to systematically assess model performance under varying conditions. This pipeline enables controlled comparison between traditional Bag-of-Words (BoW) baselines and transformer-based architectures, while explicitly examining robustness to ASR noise, codeword masking,
and adversarial message formulations. Beyond severity classification, the framework also evaluates downstream structured information extraction from transcribed maritime messages. The overall methodology is summarized in Fig. 1, which outlines the pipeline from synthetic message generation and noisy ASR processing to model training, robustness evaluation, and downstream information extraction.

### *A. SYNTHETIC DATA GENERATION PIPELINE*

The foundation of our evaluation lies in a carefully constructed synthetic dataset generated using GPT-4o-mini via the OpenAI Chat Completions API, designed to reflect the diversity of maritime radio communications across severity levels, communication styles, and scenario types. The dataset contains 1,872 messages balanced across four severity classes (Distress, Urgency, Safety, and Routine), three communication styles (formal, informal, and third_party), and 12 scenario types (*Water Ingress, Engine Failure, Fire Smoke, Collision*, etc.). Each sample includes structured metadata, including codeword-related fields such as “has codeword”, “codeword”, and “text masked”, as well as operational fields such as *Vessel Name, Call Sign, Maritime Mobile Service Identity (MMSI), Position, Persons on Board (POB), and Nature of Incident*.

The generation process followed a structured JSON-based prompting pipeline: for each sample, the generation call specified the target severity label, communication style, scenario type, and codeword condition, and the model returned a JSON object containing the message text together with associated metadata fields. Outputs were parsed, validated against the predefined schema, and discarded if they failed basic validation. Codeword fields were additionally reconciled programmatically by scanning the generated text for standard GMDSS codewords. The reference labels and metadata fields used in the experiments were not produced by applying the evaluated LLM extraction prompt to existing messages. Rather, they were part of the controlled generation design and were stored during the data-generation process. The origin of these reference fields and an independent consistency audit under the ASR-High condition are detailed in the Supplementary Material, Section S2.

Codeword occurrence was intentionally varied across the non-routine classes, while Routine messages contained no codewords by design. The codewords correspond to the standard GMDSS signal words MAYDAY, PAN-PAN, and SECURITE, and their presence or masking was controlled by the scenario metadata.

Representative examples are shown in Fig. 2, which illustrates how severity, communication style, scenario context, and codeword usage vary across the synthetic messages.

In the masked-text variant, standardized GMDSS codewords were replaced to support controlled evaluation of codeword dependence. This design enables systematic comparison across labels, styles, and masking conditions within a balanced experimental setting.

(a) Urgency | Formal | Grounding | PAN PAN

*"PAN PAN, PAN PAN, this is MV Ocean Voyager, call sign ABC1234, MMSI 123456789. We are grounded at position 35 degrees 42.5 minutes North, 139 degrees 30.2 minutes East. Weather is moderate with easterly winds of 10 knots and a slight swell. There are 12 persons on board. We require assistance to refloat the vessel. No injuries reported. Over."*

(b) Routine | Informal | Tow request | No codeword

*"Hey there, this is Ocean Breeze. We are about 15 miles east of Point Loma and having engine trouble. The sea is calm and there is no immediate danger, but we may need a tow back to port because we are not making much headway. Can anyone assist? Over."*

(c) Safety | Third-party | Steering failure | SECURITE

*"SECURITE, SECURITE. Relay from coastal station Channel Watch regarding fishing vessel Ocean's Bounty at position 34 degrees 12.5 minutes North, 118 degrees 28.9 minutes West. The vessel has experienced a steering failure, has lost maneuverability, and is currently adrift. Five persons are on board and no injuries have been reported. Nearby traffic is advised to keep clear and assist if able. Over."*

(d) Distress | Formal | Flooding | MAYDAY

*"MAYDAY, MAYDAY, MAYDAY. This is fishing vessel Ocean Hunter, call sign WXYZ123, MMSI 123456789, at position 37 degrees 45 minutes North, 122 degrees 30 minutes West. We are taking on water after striking a submerged object and require immediate assistance. Pumps are not keeping up. Six persons are on board. Over."*

**FIGURE 2. Representative synthetic message excerpts illustrating variation in severity, style, scenario, and codeword usage.**

### B. AUDIO SIMULATION AND NOISE INJECTION

To approximate degraded VHF transmission conditions in a controlled setting, we implemented a multi-stage audio processing pipeline. Clean text messages were first converted to speech using Coqui TTS [21] VITS with 109 diverse speakers with parameters optimized for maritime radio communication clarity. The generated audio files were then subjected to two levels of VHF channel noise (Medium=12dB, High=6dB SNR) using professional radio simulation tools. The noise profiles were empirically derived from actual maritime VHF recordings to ensure realistic spectral characteristics and temporal artifacts. Finally, the corrupted audio was transcribed back to text using Whisper ASR [22], introducing recognition errors that increase with noise intensity (average WER of 29.6% at Medium, and 36.2% at High noise levels).

Representative transcript examples are shown in Fig. 3, highlighting how increasing noise severity degrades transcription quality and introduces realistic ASR errors.

Clean Message

*"PAN PAN, PAN PAN, this is MV Ocean Voyager, call sign ABC1234, MMSI 123456789. We are grounded at position 35 degrees 42.5 minutes North, 139 degrees 30.2 minutes East. Weather is moderate with easterly winds of 10 knots and a slight swell. There are 12 persons on board. We require assistance to refloat the vessel. No injuries reported. Over."*

ASR Medium

*"Pan Pan. Pan Pan. This is Motive Solution Voyager, Colesign ABC-34, MMS-I-123,456,789. We are currently grounded at position 35 degrees 42.5 minutes north, 139 degrees 30.2 minutes east, weather conditions are moderate with 10 knots of wind from the east and a slight swell. There are 12 persons on board. We require assistance to reflow our vessel. No injuries reported at this time. Over.*

ASR High

*"PAM PAM. PAM PAM. This is my Diversal Ocean Voyager, call sign ABC-134-MMS-I-123,456,789. We are currently grounded at position 35 degrees, 42.5 minutes north, 139 degrees 30.2 minutes east, weather conditions are moderate with 10 knots of wind from the east and the slight snow. There are 12 persons on board, who require assistance to reflow our vessel. No injuries supported at this time. Over."*

**FIGURE 3. Examples of noise-corrupted ASR transcripts.**

### C. CLASSIFICATION MODELS AND TRAINING

Two distinct trainable modeling paradigms were implemented for comparative analysis, together with a rule-based GMDSS keyword spotter used as a zero-complexity reference baseline. The rule-based baseline assigns messages containing MAYDAY to Distress, PAN-PAN to Urgency, and SECURITE to Safety, while messages without these codewords are assigned to Routine. In the classical setting, Bag-of-Words representations were evaluated with three candidate classifiers: Logistic Regression, Linear SVM, and Naive Bayes. Based on validation performance, Logistic Regression was selected as the representative baseline model for the subsequent experiments. In the transformer setting, DistilBERT and RoBERTa-base were evaluated, and RoBERTa was selected as the representative transformer model based on validation performance. All models were evaluated using a stratified train/validation/test split (70%/15%/15%), corresponding to 1,310 training samples, 281 validation samples, and 281 test samples. The training and implementation settings of the selected trainable models are summarized in Table I. Final results are reported on the held-out test split.

TABLE I
TRAINING AND IMPLEMENTATION SETTINGS OF THE SELECTED MODELS

| Model Family | Selected Model | Key Training/Implementation Settings |
| --- | --- | --- |
| BoW Baseline | Logistic Regression | TF-IDF representation; max_iter=1000, class weight='balanced', random state=42, C=1.0 |
| Transformer | RoBERTa | Optimizer: AdamW; learning rate: 2e-5; batch size: 16; epochs: 3 |

### D. EXPERIMENTAL DESIGN

The evaluation framework comprises three classification-focused experiments, complemented by a downstream structured extraction evaluation, designed to probe different aspects of model robustness:

1. Baseline and ASR Robustness Test: Measures clean-text performance and performance degradation on ASR-transcribed messages under medium and high noise conditions using the original message labels. This evaluates basic robustness to transmission artifacts and recognition errors.
2. Adversarial Scenario Test: Introduces challenging cases including negations (e.g., “no distress”), false alarms, and procedural exercises. These test the models’ ability to interpret contextual cues beyond surface-level keyword matching.
3. Keyword Ablation Test: Assesses dependence on GMDSS codewords by comparing performance on original versus masked messages (where keywords are replaced with placeholders). This quantifies the models’ ability to utilize non-keyword contextual information.

In addition to severity classification, the framework evaluates structured information extraction from maritime messages. The extracted fields include vessel name, call sign, MMSI, position, persons on board (POB), and nature of incident. Extraction performance is assessed using field-level accuracy, with Regex-based extraction evaluated across noise conditions and compared against GPT-4o-mini-based extraction on noisy ASR transcripts.

Performance metrics include accuracy and Macro-F1 as the primary aggregate metrics, supplemented by class-wise precision, recall, and F1 analysis where relevant, with particular emphasis on Distress recall given its critical safety implications. For the ASR robustness analysis, we also report Expected Calibration Error (ECE) to assess whether model confidence scores remain reliable under noisy transcription conditions; lower ECE indicates better calibration.

## IV. RESULTS

Our experimental evaluation reveals critical insights into the comparative performance of BoW and transformer models across various operational conditions. The following subsections present detailed findings on baseline performance, noise resilience, adversarial robustness, and contextual understanding, providing a comprehensive assessment of each model’s strengths and limitations in maritime distress signal classification.

### A. BASELINE PERFORMANCE ON CLEAN DATA

The selected BoW baseline (Logistic Regression) and RoBERTa exhibit broadly comparable performance on clean text, with only a small difference between the two trainable models in the ideal input setting. The corresponding aggregate results are reported in Table II. Confidence intervals were estimated using stratified bootstrap resampling of the test-set predictions with 1,000 resamples. For each resample, Accuracy and Macro-F1 were recomputed, and the 95% confidence intervals were obtained from the 2.5th and 97.5th percentiles of the resulting bootstrap distributions.

TABLE II
CLASSIFICATION PERFORMANCE ON CLEAN DATA

| Model Type | Accuracy (95% CI) | Macro-F1 (95% CI) |
|---|---|---|
| Rule-Based GMDSS Keyword Spotter | 57.3% [51.2%, 63.0%] | 0.585[0.519, 0.641] |
| Logistic Regression (BoW) | 68% [62.6%, 73.0%] | 0.674 [0.618, 0.724] |
| RoBERTa (Transformer) | 67.6% [62.7%, 73.0%] | 0.670 [0.616, 0.720] |

Both trainable models outperform the rule-based GMDSS keyword spotter, indicating that learned lexical and contextual patterns add useful information beyond direct codeword matching. The overlapping confidence intervals for Logistic Regression and RoBERTa confirm that the clean-text performance difference between the two trainable models is small.

Class-wise analysis provides additional insight beyond the aggregate metrics. As shown in Table III, the two selected models exhibit a similar pattern across classes: Distress is the most reliably recognized category, whereas Urgency remains the most challenging. This suggests that severe emergency language is easier to identify, while intermediate-severity communications are harder to separate because their linguistic boundaries are less distinct.

TABLE III
CLASS-WISE PERFORMANCE OF THE SELECTED MODELS ON CLEAN TEXT

| Class | Logistic Regression Macro-F1 | RoBERTa Macro-F1 | Logistic Regression Recall | RoBERTa Recall |
|---|---|---|---|---|
| Routine | 0.656 | 0.7 | 0.9 | 0.886 |
| Safety | 0.655 | 0.643 | 0.521 | 0.507 |
| Urgency | 0.566 | 0.574 | 0.457 | 0.543 |
| Distress | 0.819 | 0.8 | 0.843 | 0.8 |

Error analysis on the clean test set provides a more detailed view of the aggregate results. The dominant confusion pattern in both models involves Safety and Urgency drifting toward Routine, while Urgency is also occasionally confused with Distress. RoBERTa shows a modest reduction in the most frequent Safety-to-Routine and Urgency-to-Routine errors relative to the BoW baseline, suggesting a somewhat better ability to separate lower-severity operational messages from genuinely urgent communication. Distress, by contrast, is rarely misclassified when an explicit codeword is present.

### B. ROBUSTNESS TO ASR NOISE

The robustness gap between the selected BOW baseline and RoBERTa becomes evident once clean text is replaced by noisy ASR transcripts. Although the two models perform at a broadly comparable level on clean inputs, their behavior diverges substantially under transcription corruption.

This pattern is illustrated in Table IV, which reports the model predictions for the representative message shown in Fig. 3 across its clean, ASR Medium, and ASR High versions. The values in parentheses denote the corresponding softmax confidence scores. While both models classify the clean message correctly, only RoBERTa preserves the correct Urgency label under severe ASR corruption, whereas logistic regression shifts to an incorrect Routine prediction with substantially lower confidence.

TABLE IV
EXAMPLE MODEL PREDICTIONS FOR THE MESSAGE IN FIG. 3 ACROSS ASR CONDITIONS

| | Ground Truth | Logistic Regression | RoBERTa |
|---|---|---|---|
| Clean Message | | Urgency (0.94) | Urgency (0.97) |
| ASR Medium | Urgency | Urgency (0.93) | Urgency (0.96) |
| ASR High | | Routine (0.39) | Urgency (0.92) |

As illustrated in Fig. 4, performance decreases for both models as noise increases, but the degradation is markedly steeper for Logistic Regression, whereas RoBERTa preserves stronger performance under both medium- and high-noise conditions.

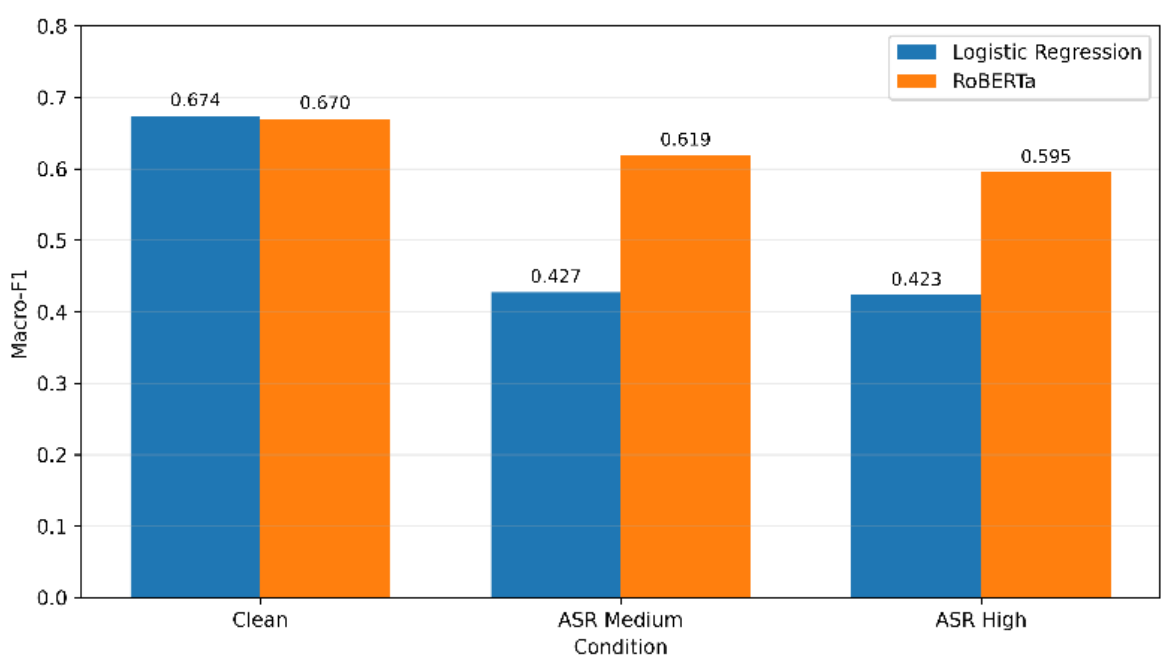

**FIGURE 4. Macro-F1 of the selected models under clean, ASR Medium, and ASR High conditions.**

This contrast is further summarized in Table V, which reports the degradation from clean text to ASR High. The BoW baseline undergoes a large relative reduction in Macro-F1, while RoBERTa shows only a modest drop. These results indicate that transformer-based modeling provides a more graceful degradation profile under ASR errors induced by simulated VHF noise, which is especially important in maritime settings where poor channel quality is common rather than exceptional.

TABLE V
PERFORMANCE DEGRADATION FROM CLEAN TEXT TO ASR HIGH

| Model | Clean Macro-F1 (95% CI) | ASR High Macro-F1 (95% CI) | Absolute Drop | Relative Drop |
|---|---|---|---|---|
| Logistic Regression | 0.674 [0.618, 0.724] | 0.423 [0.374,0.469] | -0.251 | ≈37% |
| RoBERTa | 0.670 [0.616, 0.720] | 0.595 [0.531,0.651] | -0.075 | ≈11% |

To provide statistical support for the robustness comparison, Table VI reports Macro-F1 confidence intervals, Expected Calibration Error (ECE), and paired McNemar tests for the ASR-degraded conditions. ECE is included to evaluate whether the models' confidence scores remain well calibrated under noisy ASR conditions. The results show that RoBERTa maintains higher Macro-F1 than Logistic Regression under both ASR Medium and ASR High, with statistically significant paired differences on the same test examples. RoBERTa also shows lower ECE under both noise levels, indicating better-calibrated confidence scores in the ASR-degraded setting.

TABLE VI
STATISTICAL SUMMARY OF ASR ROBUSTNESS

| Condition | Model | Macro-F1 (95% CI) | ECE | McNemar p-value vs. Logistic Regression |
|---|---|---|---|---|
| ASR Medium | Logistic Regression | 0.427 [0.375, 0.476] | 0.243 | Baseline |
| ASR Medium | RoBERTa | 0.619 [0.556, 0.675] | 0.072 | 6.78e-06 |
| ASR High | Logistic Regression | 0.423 [0.374, 0.469] | 0.2358 | Baseline |
| ASR High | RoBERTa | 0.595 [0.531, 0.651] | 0.0679 | 1.87e-04 |

To address the computational feasibility of the evaluated severity-classification models, per-message inference latency was measured for the trainable classifiers. The results are summarized in Table VII.

TABLE VII
CLASSIFICATION-STAGE INFERENCE LATENCY

| Model | ASR Medium latency, ms | ASR High Latency, ms | Mean latency, ms |
|---|---|---|---|
| Logistic Regression | 0.16 | 0.16 | 0.16 |
| RoBERTa | 6.24 | 6.43 | 6.34 |

Logistic Regression provides sub-millisecond inference, whereas RoBERTa remains within a few milliseconds per message. These results indicate that the severity-classification stage is computationally lightweight enough for low-latency message-level screening.

Class-wise analysis further clarifies the robustness gap between the two models. Fig. 5 and Table VIII show that Distress remains the most reliably recognized class, whereas Urgency is consistently the most difficult.

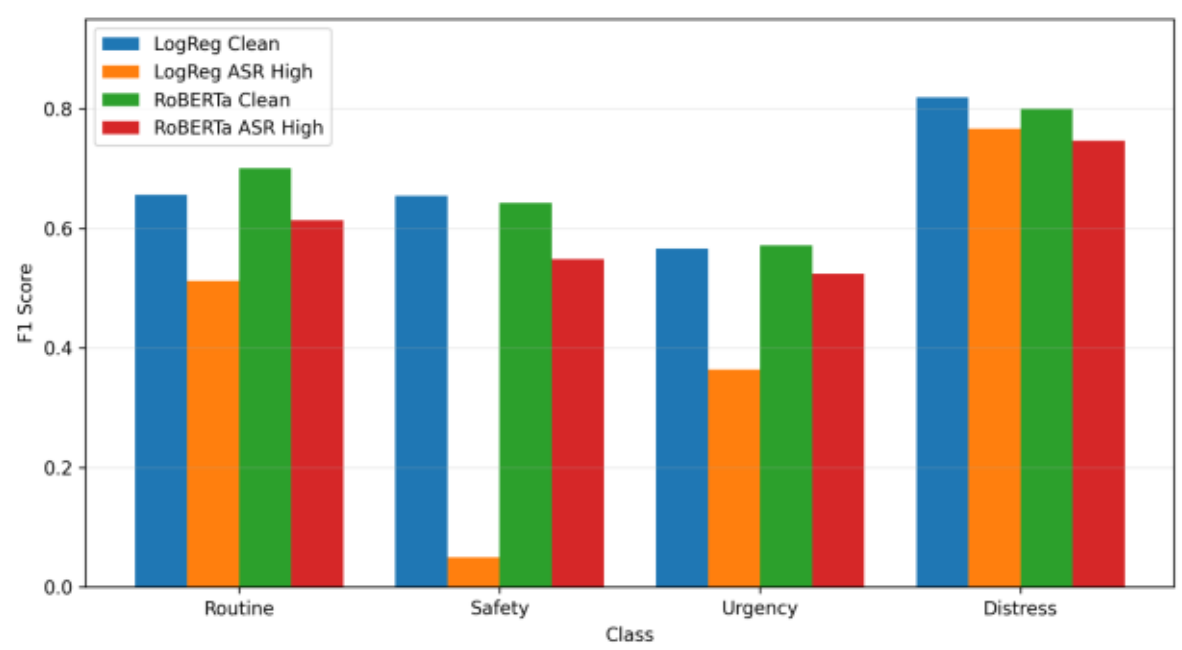

**FIGURE 5.** **Class-wise F1 of the selected models under clean and ASR High conditions.**

The largest class-wise degradation is observed for Safety in the BoW baseline, while RoBERTa maintains substantially more stable performance across all classes. These results indicate that the transformer advantage is not limited to aggregate metrics, but also extends to more robust behavior in safety-relevant categories under ASR corruption.

TABLE VIII
CLASS-WISE F1 DEGRADATION FROM CLEAN TEXT TO ASR HIGH

| Class | Logistic Regression ΔF1 (Clean-ASR High) | RoBERTa ΔF1 (Clean-ASR High) |
|---|---|---|
| Routine | 0.14 | 0.08 |
| Safety | 0.60 | 0.09 |
| Urgency | 0.20 | 0.05 |
| Distress | 0.05 | 0.05 |

A similar robustness pattern is observed across message styles under ASR High noise, where RoBERTa consistently outperforms the BoW baseline across formal, informal, and third-party communications.

These findings show that the transformer advantage under ASR corruption holds both at the aggregate and class-wise levels, motivating further analysis of model behavior under codeword masking and adversarial formulations.

### C. CODEWORD DEPENDENCE AND MASKING ROBUSTNESS

To assess how strongly the models rely on explicit GMDSS signal words, we evaluated classification performance under three masking conditions: Setting A, in which training and testing were performed on the original text; Setting B, in which codewords were masked during both training and testing; and Setting C, in which the models were trained on the original text but tested on masked text. In this experiment, the standard codewords MAYDAY, PAN-PAN, and SECURITE were replaced with the token [SIGNAL], allowing direct evaluation of how much performance depends on these lexical anchors.

The results show that both selected models are highly dependent on explicit codewords when they are available. As summarized in Fig. 6, Setting A yields large positive codeword-dependency gaps for both Logistic Regression and RoBERTa, indicating that performance is much higher on messages containing codewords than on messages without them. This suggests that, under standard conditions, both architectures rely heavily on the presence of explicit protocol terms rather than purely contextual evidence.

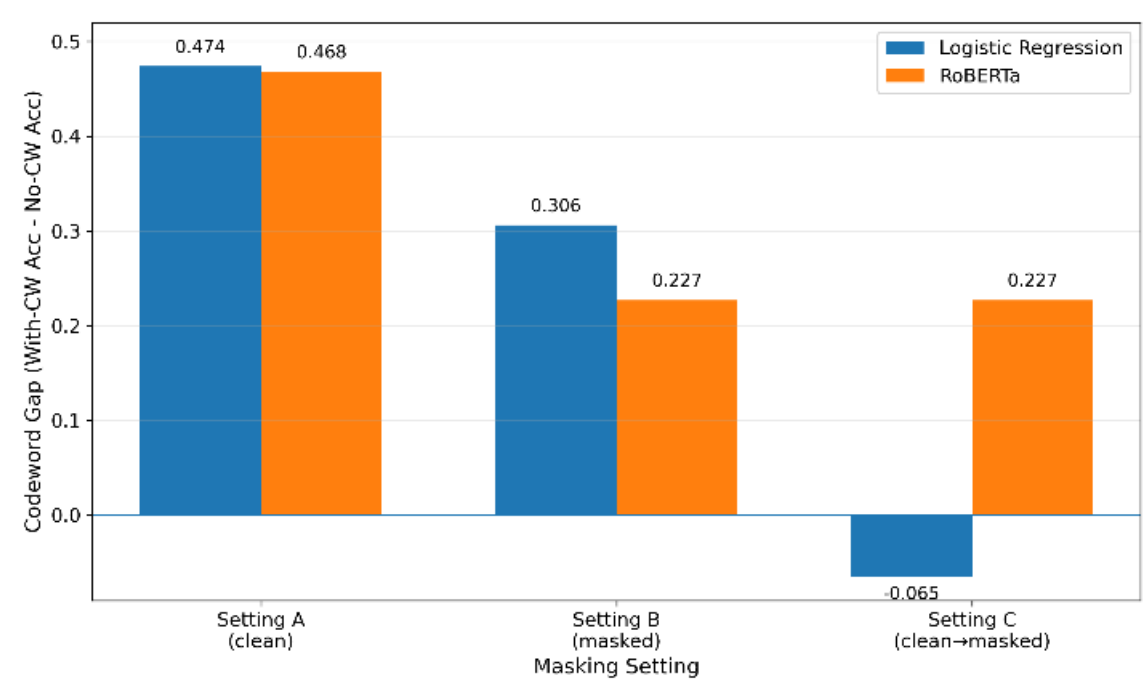

**FIGURE 6. Codeword-dependency gap across masking settings.**

Masking the codewords at both training and test time reduces overall performance, but it also reduces the codeword gap. This suggests that forcing the models to operate without explicit lexical anchors encourages greater use of contextual information, although the resulting performance remains below the original clean-text setting. RoBERTa retains an advantage under this masked setting, indicating a better ability to recover class information from the remaining message content.

The most revealing condition is Setting C, where the models are trained on the original text but tested on masked messages. Under this cross-setting evaluation, Logistic Regression degrades substantially relative to the fully masked condition, with both overall accuracy and Macro-F1 decreasing further. In addition, its codeword gap becomes negative, indicating that the model performs worse on messages that originally contained codewords than on messages that did not. By contrast, RoBERTa shows no additional degradation relative to Setting B. This contrast suggests that the transformer model transfers more reliably when explicit codewords are removed at inference time, whereas the BoW baseline is more tightly coupled to lexical patterns observed during training.

The masking experiment shows that current models do not yet achieve robust codeword-independent classification. However, the results also demonstrate that RoBERTa provides stronger fallback behavior when standard distress keywords are absent or obscured, which is particularly relevant in real maritime communication where protocol adherence may be incomplete and ASR corruption may distort critical codewords.

### D. RESILIENCE TO ADVERSARIAL TRAPS

To examine robustness beyond standard classification settings, we evaluated the selected models on a small diagnostic set of 15 hand-crafted adversarial messages designed to challenge contextual understanding. The trap set covered five categories: negation, drill, relay, ambiguous, and real_no_codeword cases. Unlike the masking experiment, which tests dependence on explicit signal words, this evaluation probes whether the models can distinguish genuine emergencies from misleading or non-operational formulations. Because of its limited size, this trap set is intended as an initial stress test rather than an exhaustive adversarial benchmark.

The adversarial results show that both models remain vulnerable in this diagnostic setting, although RoBERTa achieves higher overall robustness. As summarized in Table IX, RoBERTa attains higher overall accuracy than the BoW baseline on the trap set, indicating a better ability to handle challenging formulations that cannot be resolved through simple keyword matching alone. However, the absolute performance of both models remains limited, suggesting that adversarially phrased maritime messages continue to pose a substantial challenge. The small sample size prevents strong generalization across trap categories, so these results should be interpreted as evidence of important failure modes that require broader future benchmarking.

TABLE IX
ACCURACY OF THE SELECTED MODELS ACROSS ADVERSARIAL TRAP TYPES

| Trap Type | n | Logistic Regression | RoBERTa |
|---|---|---|---|
| Negation | 4 | 0.25 | 0.25 |
| Drill | 3 | 0 | 0 |
| Relay | 2 | 0.5 | 0.5 |
| Ambiguous | 3 | 0.333 | 0.667 |
| Real_no_codeword | 3 | 0.333 | 0.667 |
| Overall | 15 | 0.267 | 0.4 |

Performance varies considerably across trap types. The clearest advantage for RoBERTa appears in the ambiguous and real_no_codeword categories, where it outperforms the BoW baseline by a visible margin. By contrast, both models perform identically on negation and relay cases, and both fail completely on drill messages. This last result is especially important operationally, because it implies that training or exercise traffic could be incorrectly treated as real emergencies if such systems were deployed without additional safeguards.

The qualitative examples reinforce this interpretation. In several cases where BoW was misled by surface lexical cues, RoBERTa correctly identified the intended class, including routine operational phrasing and drill-related language. At the same time, the zero accuracy on drill messages shows that even the stronger model is not sufficiently reliable for unsupervised use in adversarial or procedurally marked communication scenarios. This drill-message failure is operationally important because training and exercise traffic may contain emergency-like terminology while not representing a real incident, and future systems should explicitly model drill markers, exercise context, and human review safeguards.

### *E. STRUCTURED INFORMATION EXTRACTION*

Beyond severity classification, we evaluated the extraction of operational fields from maritime messages under clean and noisy ASR conditions. The extracted fields included vessel name, call sign, MMSI, position, persons on board (POB), and nature of incident, with Regex-based extraction evaluated across all noise levels and a head-to-head comparison between Regex and zero-shot GPT-4o-mini-based extraction. The reference fields used for extraction evaluation were metadata fields defined and stored during dataset construction. GPT-4o-mini-based extraction used JSON-mode output with temperature set to 0; the full extraction prompt, output constraints, invalid-output handling, field mapping, and matching rules are provided in the Supplementary Material, Section S1. Additional details on the origin of the reference fields and the ASR-High consistency audit are provided in the Supplementary Material, Section S2. The corresponding comparison results are summarized in Figs. 7 and 8.

As shown in Fig. 7, Regex extraction performance is highly field-dependent and degrades sharply under ASR corruption. Vessel is the most extractable field on clean text, but its accuracy still declines substantially at high noise. By contrast, Call Sign and MMSI are especially brittle, because ASR corruption severely disrupts alphanumeric strings. POB is comparatively more stable across noise levels, whereas Position and Nature show moderate but consistent degradation as noise increases.

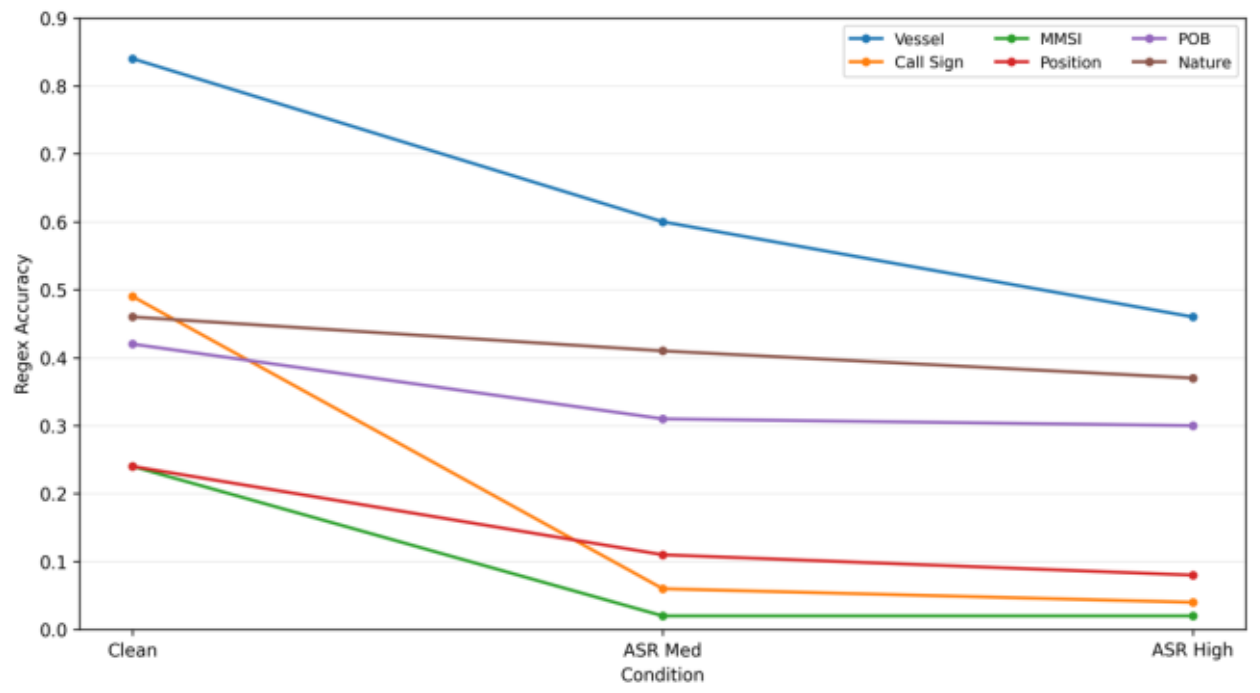

**FIGURE 7. Regex extraction accuracy across clean and noisy ASR conditions.**

The comparison at ASR High in Fig. 8 shows a clear advantage for GPT-4o-mini-based extraction over Regex across all evaluated fields. The largest relative improvements occur for MMSI, Call Sign, and Position, where rule-based extraction becomes nearly non-functional under noisy transcription. GPT-4o-mini also provides strong gains for POB and Nature, indicating better recovery of semantically expressed information even when surface forms are corrupted. Although Vessel remains the easiest field for Regex, GPT-4o-mini-based extraction still improves it substantially under high noise.

The extraction results show that structured information recovery from maritime ASR transcripts cannot rely on Regex alone under the noisy conditions examined here. Rule-based extraction remains usable for some relatively stable fields, but stronger performance is obtained with zero-shot LLM-based extraction in this controlled benchmark, especially for alphanumeric identifiers and semantically variable free-text fields.

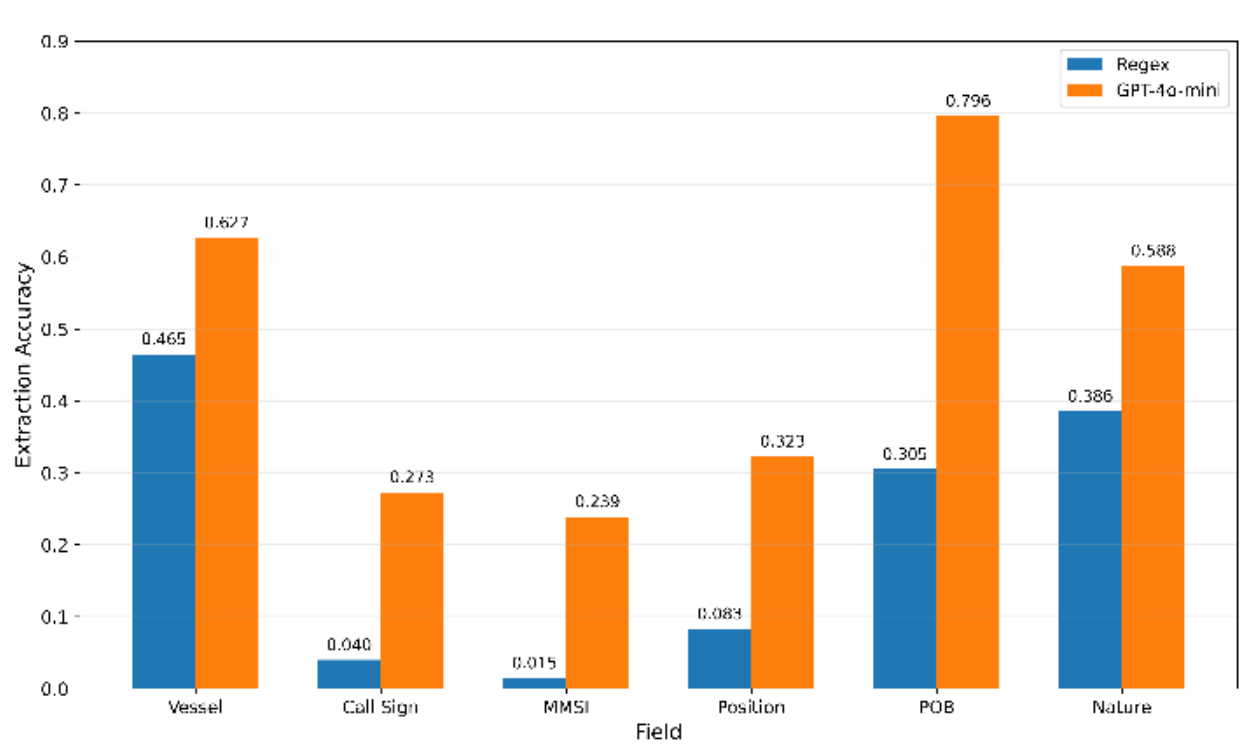

**FIGURE 8.** **Regex vs. GPT-4o-mini-based extraction accuracy at ASR High.**

Table X presents an illustrative extraction example from a noisy ASR High transcript:

> *"PAM PAM. PAM PAM. This is my Diversal Ocean Voyager, call sign ABC-134-MMS-I-123,456,789. We are currently grounded at position 35 degrees, 42.5 minutes north, 139 degrees 30.2 minutes east, weather conditions are moderate with 10 knots of wind from the east and the slight snow. There are 12 persons on board, who require assistance to reflow our vessel. No injuries supported at this time. Over."*

TABLE X
ILLUSTRATIVE GPT-4O-MINI-BASED EXTRACTION OUTPUT FOR A NOISY ASR HIGH TRANSCRIPT

| Field | Ground Truth | GPT-4o-mini-based Information Extraction |
| --- | --- | --- |
| Vessel | MV OCEAN Voyager | Diversal Ocean Voyager |
| Call Sign | ABC1234 | ABC-134-MMS-I-123 |
| MMSI | 123456789 | 123456789 |
| Position | 35 degrees 42.5 minutes North, 139 degrees 30.2 minutes East | 35 degrees, 42.5 minutes north, 139 degrees 30.2 minutes east |
| POB | 12 | 12 |
| Incident | grounding | Currently grounded, no injuries reported. |

As shown in Table X, despite severe transcript corruption, GPT-4o-mini-based extraction successfully recovers most of the operationally critical fields, including MMSI, position, POB, and incident type, while only minor errors remain in some surface-form details such as vessel name and call sign. The extracted output is therefore largely consistent with the reference metadata fields in this example, although some surface-form errors remain. In contrast, Regex failed on this example because the noisy transcript disrupted the lexical and formatting patterns required for reliable rule-based matching, further highlighting the advantage of LLM-based extraction under ASR noise induced by simulated VHF degradation.

## V. DISCUSSION

The results show that the main advantage of the transformer-based approach does not lie in clean-text classification, where performance remains broadly comparable to strong Bag-of-Words baselines, but in robustness under simulated ASR degradation. Under ASR corruption, RoBERTa exhibits a much smaller drop in Macro-F1 than Logistic Regression, indicating that contextual modeling becomes more valuable once lexical surface forms are distorted by transcription noise. This finding is important for maritime communication, where poor channel quality can be a common operating condition rather than an exceptional case. This interpretation is reinforced by the class-wise analysis: Distress remains the most stable category under noise, whereas Safety shows a severe collapse for the BoW baseline under medium ASR noise, representing a critical failure mode for a safety-critical system. The calibration results further support this interpretation: RoBERTa shows lower ECE than Logistic Regression under both ASR Medium and ASR High, suggesting that its confidence scores are more reliable under transcription noise. Additional cross-validation, learning-curve, and ROC-AUC analyses are provided in the Supplementary Material, Sections S3 and S4.

The masking experiments further clarify the nature of this robustness advantage. Both selected models rely strongly on explicit GMDSS codewords when they are available, which means that current systems still exploit protocol-specific lexical anchors rather than fully semantic representations of maritime emergencies. However, the two models differ markedly when these anchors are removed. RoBERTa maintains stable performance across the masked and cross-setting evaluations, whereas Logistic Regression deteriorates further and even shows a negative codeword gap in the clean-to-masked transfer setting. This suggests that the transformer model provides stronger fallback behavior when standard codewords are absent, omitted, or corrupted in transmission. The masking results are also consistent with the ASR analysis, which shows that although 72.9% of codewords remain recoverable at high noise, 27.1% are lost, directly contributing to downstream classification failures.

The adversarial evaluation adds an important operational qualification. Although RoBERTa outperforms the BoW baseline overall, the present trap set is small and should be treated as diagnostic rather than exhaustive. Both models remain vulnerable to deceptive and procedurally marked messages, and both fail completely on drill traffic. This is a critical concern for any future deployment scenario, because an automated system that cannot distinguish exercises from genuine emergencies may generate unsafe false alarms. Accordingly, the present results support the use of such models as decision-support tools rather than fully autonomous detectors. Human oversight, especially for borderline, procedurally unusual, or drill-related messages, remains essential.

The structured extraction results extend the same robustness story to downstream information recovery. Regex-based extraction degrades sharply under ASR noise, especially for alphanumeric and format-sensitive fields such as MMSI and Call Sign. GPT-4o-mini-based extraction, by contrast, outperforms Regex on every evaluated field at ASR High within this benchmark, with especially large gains for MMSI, Call Sign, Position, and POB. This suggests that in practical maritime pipelines, robust classification alone is insufficient; extraction methods must also tolerate phonetic distortion, paraphrasing, and partial corruption of structured content. In this controlled setting, LLM-based extraction appears more suitable than purely rule-based parsing, while real-world claims require independent validation on operational traffic and human-verified annotations. An independent audit using a second model family and human adjudication, reported in the Supplementary Material, Section S2, supports the validity of these reference fields under the noisy conditions examined here.

Several limitations should be acknowledged. First, the study relies on a balanced synthetic dataset generated to support controlled experimentation, rather than on real operational VHF traffic. This enables systematic comparison but may not capture the full linguistic and distributional variability of real-world maritime communication. Second, the ASR analysis is based on simulated noisy conditions, which approximate but do not exhaust the acoustic complexity of real channels. In addition, the reported results may depend on the selected TTS voices, noise profiles, signal-to-noise ratios, temporal degradation patterns, and transcription model. Third, the adversarial trap set is intentionally focused and small, so the corresponding results should be interpreted as diagnostic rather than exhaustive. These limitations point to several natural directions for future work, including evaluation on real maritime recordings, construction of a larger standardized benchmark for synthetic and real maritime communication data, broader adversarial testing with 30-50 or more examples per trap category, systematic sensitivity analysis across TTS voices, noise profiles, temporal deletion, and ASR pipelines, comparison with alternative data-construction and augmentation strategies, such as human-authored scenarios, rule-based templates, paraphrase augmentation, smaller domain-adapted generators, and real-recording augmentation, and end-to-end audio or multimodal models that reduce cascading ASR errors.

The study suggests that robust maritime communication intelligence requires more than strong clean-text classification. The central challenge is resilience under degraded, incomplete, and non-standard communication, and in this respect transformer-based classification and LLM-based extraction provide a more robust foundation than lexical baselines and rule-based parsing within the controlled setting evaluated here.

## VI. CONCLUSION

This study presented SeaAlert, a controlled evaluation framework for maritime distress communication analysis that jointly addresses severity classification and structured information extraction from noisy ASR transcripts. Using a balanced synthetic dataset of 1,872 maritime radio messages, we compared Bag-of-Words baselines and transformer-based classifiers under clean-text, noisy-ASR, codeword-masking, and diagnostic adversarial conditions, and also evaluated downstream structured extraction using Regex and GPT-4o-mini. The results show that while clean-text performance is broadly comparable, RoBERTa is less degraded by ASR corruption, degrades more gracefully when explicit codewords are removed, and achieves higher accuracy on the small diagnostic adversarial trap set. At the same time, the experiments show that neither classifier is fully robust to deceptive procedural traffic, particularly drill messages.

For structured information extraction, the study shows that Regex-based methods are insufficient under the simulated noisy-ASR conditions examined here, especially for alphanumeric and format-sensitive fields. The practical implication is especially clear for fields such as MMSI and Call Sign, where Regex becomes nearly non-functional under ASR corruption, whereas GPT-4o-mini-based extraction outperforms Regex across all evaluated fields in this controlled benchmark. This indicates that LLM-based extraction is a promising approach within the simulated noisy-audio setting examined here, but real-world deployment requires independent validation on operational maritime traffic and human-verified annotations. Taken together, these findings suggest that future maritime communication systems should combine robust severity assessment with resilient information extraction, while preserving human oversight for ambiguous or procedurally complex cases. Future work should validate these conclusions on real maritime traffic, broaden adversarial and sensitivity testing, develop standardized benchmarks for maritime communication data, compare alternative generation and augmentation strategies, and investigate direct audio or multimodal architectures that reduce codeword dependence and limit error propagation from ASR.

# Supplementary Material

## S1. GPT-4o-mini-Based Structured Extraction Reproducibility Details

### *S1.A Model, API, and Decoding Settings*

The GPT-4o-mini-based structured extraction experiment was separate from the synthetic message-generation stage. It took noisy ASR transcripts as input and returned structured fields in JSON format. These extracted fields were then compared with reference metadata fields that had been defined and stored during dataset construction. The structured information extraction experiment used GPT-4o-mini via the OpenAI Chat Completions API. The model was applied to ASR High transcripts and instructed to return predefined maritime fields in JSON format. Table S.I summarizes the GPT-4o-mini extraction-stage configuration.

TABLE S.I
GPT-4o-mini extraction-stage configuration

| Parameter | Setting |
|---|---|
| Model | gpt-4o-mini |
| API | OpenAI Chat Completions |
| Response format | JSON object mode |
| Temperature | 0 |
| Output keys | vessel, call_sign, mmsi, position, pob, nature, codeword |

Temperature was set to 0 to minimize sampling variability under fixed prompt and model settings. JSON-mode output was enforced at the API level to reduce formatting variation and prevent prose or markdown responses.

### *S1.B Prompt Design*

**System prompt**

```
You are a maritime rescue communications AI.
Extract structured information from noisy ASR transcripts of distress calls.

Rules:
1. Handle phonetic errors (e.g., "tree" -> "3", "niner" -> "9", "m m s eye" -> "MMSI").
2. Contextually correct maritime terms (e.g., "Vesal" -> "Vessel", "may day" -> "Mayday").
3. Return ONLY a valid JSON object with keys:
   "vessel", "call_sign", "mmsi", "position", "pob", "nature", "codeword".
4. "codeword" refers to the emergency declaration (e.g., "Mayday", "Pan Pan", "Securite").
5. "nature" refers to the type of distress or incident (e.g., "fire", "sinking", "collision").
6. If a field is missing, set its value to null.
```

**User prompt template**

```
Extract info from: {asr_high_transcript}
```

### *S1.C Output Handling and Evaluation*

The model output was constrained to a valid JSON object. If a field was not mentioned in the transcript, the model was instructed to return null. Empty or non-string inputs were treated as failed extraction cases. If an API error, invalid response, or JSON parsing failure occurred, the extraction returned an empty result and the corresponding fields were scored as failed extractions.

Extracted values were compared with the reference metadata fields defined and stored during dataset construction, rather than with outputs produced by the extraction prompt. Fuzzy string matching was used, with a partial-ratio threshold of 70 for a correct extraction. Null reference fields were excluded from evaluation, and missing or empty extracted values were scored as incorrect. Table S.II summarizes the field mapping and matching rules used for extraction evaluation.

TABLE S.II
FIELD MAPPING USED FOR EXTRACTION EVALUATION

| Extracted Field | Reference Metadata Field | Matching Rule |
|---|---|---|
| vessel | vessel | Fuzzy match |
| call_sign | call_sign | Fuzzy match |
| mmsi | mmsi | Fuzzy match |
| position | location | Fuzzy match |
| pob | pob | String comparison |
| nature | nature | Fuzzy match |
| codeword | codeword | Fuzzy match |

## S2. Reference-Field Origin and ASR-High Consistency Audit

### *S2.A Reference-Field Origin*

The structured information extraction task was evaluated against metadata fields stored during dataset construction. These reference fields were not produced by applying the evaluated GPT-4o-mini extraction prompt to the messages. Rather, they were stored as part of the synthetic sample construction process and later used as reference fields for extraction evaluation.

### *S2.B Independent Consistency Audit*

To further examine reference-field reliability under the same noisy condition used for structured extraction, we conducted an independent ASR-High consistency audit. Claude Sonnet was used as an external re-annotation/checking model and extracted the same structured fields from ASR High transcripts. Its outputs were compared with the stored reference fields using the same fuzzy-matching threshold used in the extraction evaluation. A stratified sample of 48 messages was then manually adjudicated by a human reviewer. Table S.III summarizes the ASR-High consistency audit design and aggregate audit metrics.

TABLE S.III
ASR-HIGH CONSISTENCY AUDIT SUMMARY

| Audit Item | Value |
|---|---|
| Messages screened by independent model | 1,872 |
| Manually adjudicated sample | 48 messages |
| Severity-class coverage | 4 classes × 12 messages |

| Style coverage | 3 styles × 16 messages |
|---|---|
| Human-adjudicated field-level checks | 323 |
| Human-adjudicated consistency rate | 58.8% |
| Claude-human agreement on audit decisions | 86.1% |

Claude-human agreement was computed over binary field-level consistency decisions, not as a direct extraction-accuracy score.

This audit should be interpreted as a noisy-transcript consistency and recoverability check under ASR High conditions, rather than as validation on real operational traffic. Table S.IV reports the human-adjudicated field-level consistency results.

TABLE S.IV
HUMAN-ADJUDICATED FIELD-LEVEL CONSISTENCY UNDER ASR HIGH

| Field | Consistent | Total | Rate |
|---|---|---|---|
| Nature of incident | 44 | 48 | 91.7% |
| Codeword | 27 | 36 | 75.0% |
| Location | 44 | 48 | 91.7% |
| POB | 35 | 48 | 72.9% |
| Vessel | 23 | 48 | 47.9% |
| MMSI | 9 | 47 | 19.1% |
| Call sign | 8 | 48 | 16.7% |
| Overall | 190 | 323 | 58.8% |

The audit shows that consistency under ASR High noise is strongly field-dependent. Semantic and operational fields such as nature of incident, codeword, location, and POB were more consistently recoverable, whereas exact identifiers such as call sign and MMSI were much more vulnerable to ASR corruption. These findings support the manuscript's interpretation that structured extraction from noisy maritime transcripts is field-dependent and that alphanumeric identifiers require particular caution in deployment-oriented settings.

## S3. Cross-Validation and Learning-Curve Analysis

### *S3.A Purpose and Evaluation Setting*

To further assess clean-text split-level stability and data-scaling behavior, we conducted supplementary cross-validation and learning-curve analyses. These analyses were performed on the combined train and validation portion of the dataset, excluding the held-out test set to avoid data leakage. The held-out test-set results reported in the main manuscript remain the primary evaluation results.

### *S3.B Five-Fold Cross-Validation on Clean Text*

Stratified 5-fold cross-validation was performed on the combined train and validation set. This supplementary analysis evaluates clean-text split-level stability and complements, but does not replace, the held-out test-set results reported in the main manuscript. Table S.V reports the five-fold cross-validation results on clean text.

TABLE S.V
FIVE-FOLD CROSS-VALIDATION RESULTS ON CLEAN TEXT

| Fold | Logistic Regression Macro-F1 | RoBERTa Macro-F1 |
|---|---|---|
| Fold 1 | 0.6697 | 0.7235 |
| Fold 2 | 0.6996 | 0.7730 |
| Fold 3 | 0.6902 | 0.7426 |
| Fold 4 | 0.7037 | 0.7440 |
| Fold 5 | 0.7353 | 0.7370 |
| Mean | 0.6997 | 0.7440 |
| Std | 0.0213 | 0.0162 |
| 95% CI | [0.6579, 0.7415] | [0.7122, 0.7758] |

Across the five clean-text folds, RoBERTa achieved a higher mean Macro-F1 than Logistic Regression and showed slightly lower fold-to-fold variability. These results provide supplementary evidence of clean-text split-level stability across internal splits. They should be interpreted alongside, and not as a replacement for, the held-out test-set results in the main manuscript, where clean-text performance of the two trainable models is reported as broadly comparable.

### S3.C Learning-Curve Analysis

Learning curves were computed by training the models on progressively larger subsets of the training data. The purpose was to assess whether performance continued to improve substantially as more synthetic training data were added. Figure S1. Learning-curve analysis on clean-text validation data.

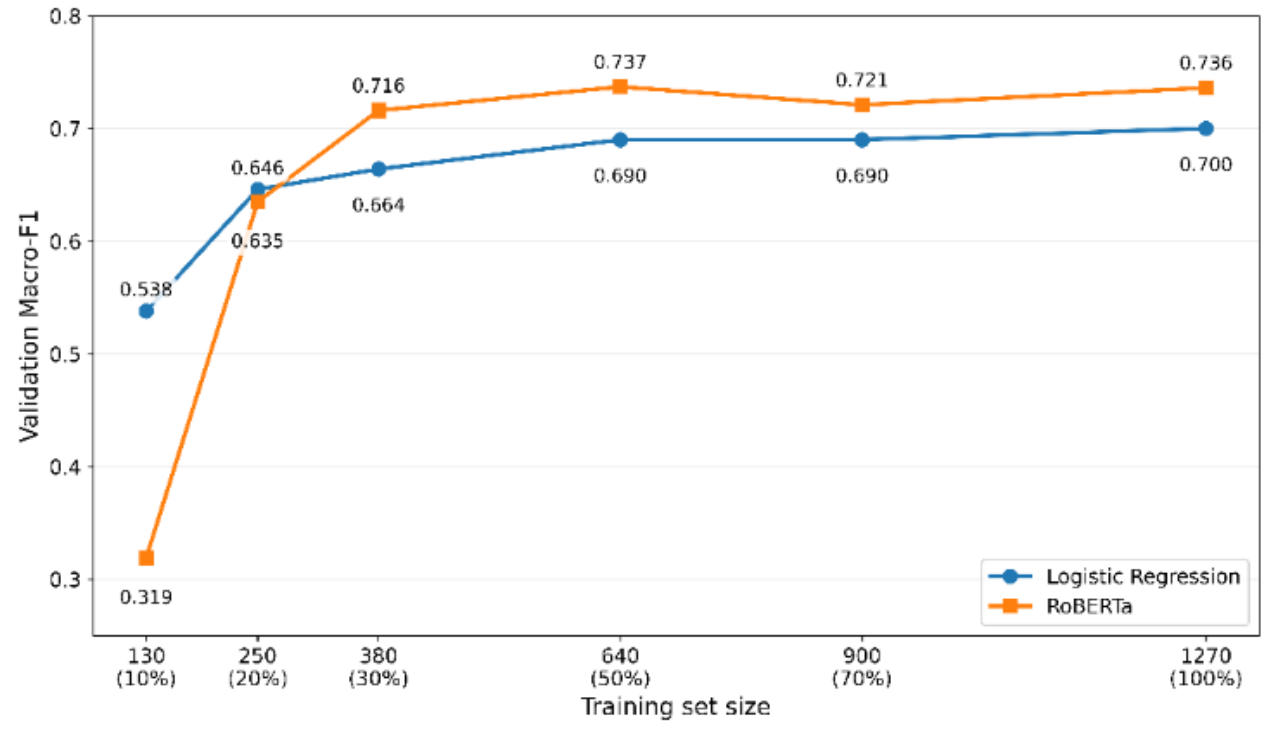

**Figure S1. Learning-curve analysis on clean-text validation data. Validation Macro-F1 is shown for Logistic Regression and RoBERTa across increasing training-set sizes. Both models improve rapidly with additional data at small training sizes and show diminishing returns beyond approximately 50% of the training data in this controlled clean-text setting.**

The learning-curve analysis suggests that most of the clean-text performance gain is obtained by approximately half of the training data. Logistic Regression approaches a validation Macro-F1 of approximately 0.70, while RoBERTa stabilizes around 0.73 to 0.74. This indicates that performance begins to plateau in the present controlled clean-text experiment. However, the final experiments used the full training set to maximize coverage across severity labels, communication styles, scenario types, and codeword conditions, and to ensure that the robustness evaluations were based on the strongest available trained models. Larger and more diverse datasets may still be needed for broader real-world generalization.

## S4. ROC-AUC Analysis

The main manuscript reports Accuracy, Macro-F1, confidence intervals, calibration, latency, and paired statistical comparisons. As an additional measure of class separability, we computed supplementary ROC-AUC values for the trainable Logistic Regression and RoBERTa classifiers under ASR Medium and ASR High conditions.

The rule-based GMDSS keyword spotter is not included in the ROC-AUC comparison because it does not produce continuous class-probability scores comparable to the trainable models. Its performance is therefore reported in the main manuscript using Accuracy and Macro-F1. Table S.VI reports aggregate AUC values under ASR Medium and ASR High conditions.

TABLE S.VI
AGGREGATE AUC UNDER ASR NOISE

| Noise Condition | Model | AUC (95% CI) |
|---|---|---|
| ASR Medium | Logistic Regression | 0.7538 [0.7160, 0.7902] |
| ASR Medium | RoBERTa | 0.8221 [0.7869, 0.8533] |
| ASR High | Logistic Regression | 0.7284 [0.6897, 0.7643] |
| ASR High | RoBERTa | 0.8042 [0.7678, 0.8370] |

RoBERTa maintains a higher aggregate AUC than Logistic Regression under both ASR Medium and ASR High conditions. This supports the Macro-F1 results reported in the main manuscript and indicates stronger class-separation behavior under noisy transcription conditions. Table S.VII reports the corresponding per-class AUC values.

TABLE S.VII
PER-CLASS AUC UNDER ASR NOISE

| Noise Condition | Class | Logistic Regression AUC | RoBERTa AUC |
|---|---|---|---|
| ASR Medium | Distress | 0.937 | 0.926 |
| ASR Medium | Routine | 0.782 | 0.850 |
| ASR Medium | Safety | 0.673 | 0.778 |
| ASR Medium | Urgency | 0.623 | 0.735 |
| ASR High | Distress | 0.932 | 0.899 |
| ASR High | Routine | 0.746 | 0.826 |
| ASR High | Safety | 0.649 | 0.761 |
| ASR High | Urgency | 0.587 | 0.731 |

The per-class AUC results show that both models achieve high AUC for Distress, reflecting the strong lexical signal associated with explicit emergency codewords. RoBERTa shows higher AUC for Routine, Safety, and Urgency under both reported ASR noise conditions, indicating better separation for classes that require more contextual interpretation beyond a single codeword cue.

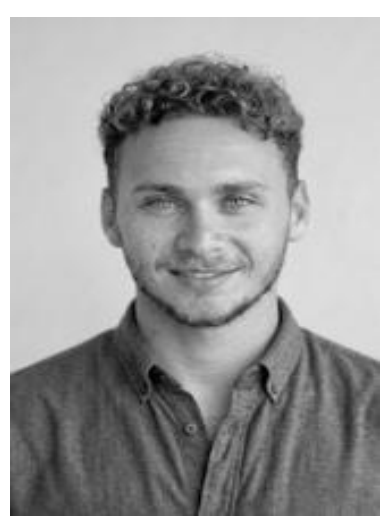

**TOMER ATIA** is currently working toward his B.Sc. degree in Computer Science, with specialization in deep learning and artificial intelligence, at the Holon Institute of Technology (HIT), Holon, Israel. He is expected to receive his degree in 2026. His research interests include computer vision, natural language processing, large language models, generative AI, and deep learning for safety-critical systems, with a particular focus on maritime applications.

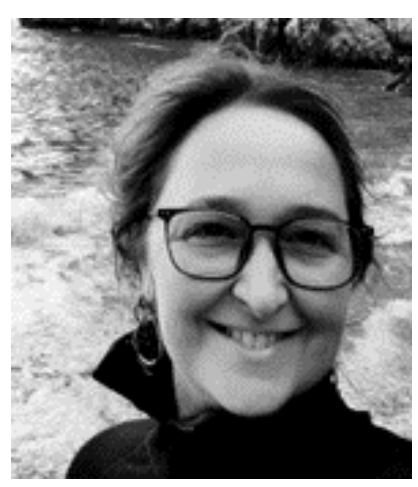

**YEHUDIT APERSTEIN** received her Ph.D. degree in applied mathematics from the Faculty of Mathematics and Computer Science of the Weizmann Institute of Science. She completed postdoctoral fellowships at Tel Aviv University and Bar-Ilan University. She is currently the Head of Afeka Interdisciplinary Center for Social Good Generative AI at Afeka Academic College of Engineering in Tel Aviv. Her research interests include robust and trustworthy artificial intelligence, explainable AI, natural language processing, generative AI, and deep learning for intelligent systems, with applications in real-world multidisciplinary domains.

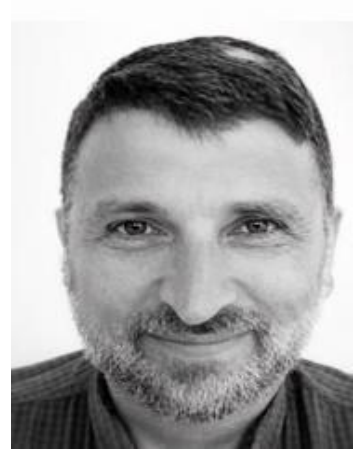

**ALEXANDER APARTSIN** received his B.Sc., M.Sc., and Ph.D. degrees in Computer Science from the Technion, Israel Institute of Technology, the Weizmann Institute of Science, and Tel Aviv University, respectively. He also completed a postdoctoral fellowship in computer science at the Weizmann Institute of Science. He brings over 30 years of industry experience in software engineering, product development, data science, and research leadership. He is currently a faculty member with the Computer Science Department, Holon Institute of Technology (HIT), Holon, Israel. His research interests include generative AI, synthetic data generation, natural language processing, computer vision, robotics, anomaly detection, and AI for data-scarce domains, including education, healthcare, and defense.